\documentclass[letterpaper, 10 pt, conference]{ieeeconf}
\IEEEoverridecommandlockouts

\usepackage{times}

\usepackage[hyphens]{url}  %
\usepackage{graphicx} %
\usepackage{caption} %

\usepackage{multicol}
\usepackage[bookmarks=true]{hyperref}

\graphicspath{{../}{./}{./figures/}}
\usepackage[utf8]{inputenc}

\usepackage{svg}
\usepackage{amsmath}
\usepackage[font=small,labelfont=bf]{caption}
\usepackage{graphicx}
\usepackage{xcolor}
\usepackage[justification=centering]{subfig}
\usepackage{amsmath}
\usepackage{amsfonts}
\usepackage{amssymb}
\usepackage{mathtools}
\usepackage{todonotes}
\let\labelindent\relax
\usepackage{enumitem}
\usepackage{comment}
\usepackage{booktabs}
\usepackage[per-mode=symbol]{siunitx}
\DeclareSIUnit\minute{min}
\usepackage{tikzpagenodes}
\usepackage[noend]{algorithm,algorithmic}
\floatname{algorithm}{Algorithm}

\usepackage{geometry}
\newcommand{\ph}[1]{{\textbf{#1:}}} %
\newcommand{\ncomment}[1]{}
\newcommand{\qcomm}[1]{{\color{red}Quest.}} %

\newcommand{\argmax}{\mathop{\mathrm{argmax}}}

\newcommand{\senseFunc}{CoverageUpdate}

\newcommand{\state}{$s = (n_q, \mu, G)$}

\usepackage{flushend}

\DeclareUnicodeCharacter{0301}{*************************************}

\begin{document}
\hypersetup{hidelinks,colorlinks=false}

\bstctlcite{IEEEexample:BSTcontrol}

\title{Adaptive Coverage Path Planning for Efficient \\ Exploration of Unknown Environments

\thanks{
*The work is partially supported by the Jet Propulsion Laboratory, California Institute of Technology, under a contract with the National Aeronautics and Space Administration (80NM0018D0004), and Defense Advanced Research Projects Agency (DARPA).}

}

\author{
    Amanda Bouman\textsuperscript{\rm 1}\thanks{\textsuperscript{\rm 1}Department of Mechanical and Civil Engineering, California Institute of Technology (e-mail: \{abouman, jwb@robotics\}.caltech.edu).},
    Joshua Ott\textsuperscript{\rm 2}\thanks{\textsuperscript{\rm 2}Department of Aeronautics and Astronautics, Stanford University (e-mail: \{joshuaott, mykel\}\!@stanford.edu).},
    Sung-Kyun Kim\textsuperscript{\rm 3}\thanks{\textsuperscript{\rm 3}NASA Jet Propulsion Laboratory, California Institute of Technology (e-mail: \{sung.kim, aliahga\}\!@jpl.nasa.gov).}, 
    Kenny Chen\textsuperscript{\rm 4}\thanks{\textsuperscript{\rm 4}Department of Electrical and Computer Engineering, University of California Los Angeles (e-mail: kennyjchen@ucla.edu).}, \\
    Mykel J. Kochenderfer\textsuperscript{\rm 2},
    Brett Lopez\textsuperscript{\rm 5}\thanks{\textsuperscript{\rm 5}Department of Mechanical and Aerospace Engineering, University of California Los Angeles (e-mail: btlopez@ucla.edu).},
    Ali-akbar Agha-mohammadi\textsuperscript{\rm 3},
    Joel Burdick\textsuperscript{\rm 1}
}

\maketitle

\begin{abstract}
We present a method for solving the coverage problem with the objective of autonomously exploring an unknown environment under mission time constraints.  
Here, the robot is tasked with planning a path over a horizon such that the accumulated area swept out by its sensor footprint is maximized. 
Because this problem exhibits a \emph{diminishing returns} property known as submodularity, we choose to formulate it as a tree-based sequential decision making process.
This formulation allows us to evaluate the effects of the robot's actions on future world coverage states, while simultaneously accounting for traversability risk and the dynamic constraints of the robot.
To quickly find near-optimal solutions, we propose an effective approximation to the coverage sensor model which adapts to the local environment.
Our method was extensively tested across various complex environments and served as the local exploration algorithm for a competing entry in the DARPA Subterranean Challenge.

\end{abstract}

\IEEEpeerreviewmaketitle

\section{Introduction}\label{sec:intro}

Consider a time-limited mission wherein a ground robot must autonomously explore an unknown environment with complex terrain. 
The robot explores by maximizing the area observed, or \emph{covered}, by a task-specific \emph{coverage sensor}. This sensor may be a thermal camera for detecting thermal signatures, an optical camera for identifying visual clues, or in our case, an omnidirectional range finder for constructing 3D environment maps. As the robot moves, the sensor footprint sweeps the environment, expanding the covered area, or more generally, the task-relevant information about the world. The problem of finding efficient and safe coverage trajectories is computationally complex \cite{sungpilgrim, heng2015efficient}-- one must consider the fact that a robot’s observation of the world affects the utility of future observations, while concurrently minimizing traversability risk.

Our proposed method quickly finds non-myopic coverage paths by rolling out future coverage observations using an effective sensor model.  Our model is carefully designed to replicate critical features of a range finder in a computationally efficient manner.
First, the model is probabilistic -- coverage probability decreases with increasing ray sparsity along the radial direction. As an effect, the density of coverage is dictated by the local environment geometry, and large topological features in the environment are quickly exposed and mapped. Second, to account for ray-surface interactions that regulate surface visibility, the coverage range, or distance at which a sensor measurement is performed, adapts to the scale of the local environment. This approach obviates the need for expensive ray-tracing operations that make forward rollout algorithms prohibitively slow for a real-time system. 

We begin by noting that the coverage task is submodular. Since the robot must understand the effects of its actions on the quality of future coverage measurements, we choose to formulate this problem as a sequential decision process. To find near-optimal trajectories at high replanning rates, we use an online forward rollout search algorithm that plans from the current world-robot state to a travel budget-defined horizon. Our method was evaluated on hardware in various environments, and served as the local planner for team CoSTAR's entry in the Final Circuit of the DARPA SubT Challenge \cite{AliNeBula21}.

\section{Related Work}\label{sec:relatedwork}
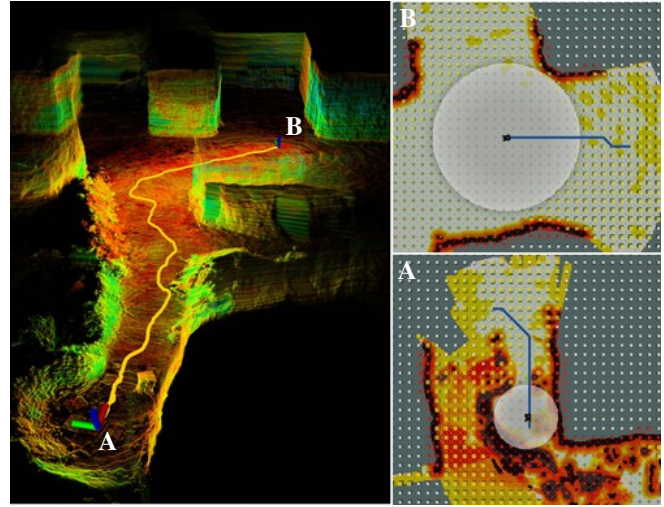
\begin{figure}[t]
\centering
    \begin{tikzpicture}
    \node[anchor=south west,inner sep=0] (image) at (0,0) {\includegraphics[width=1\columnwidth]{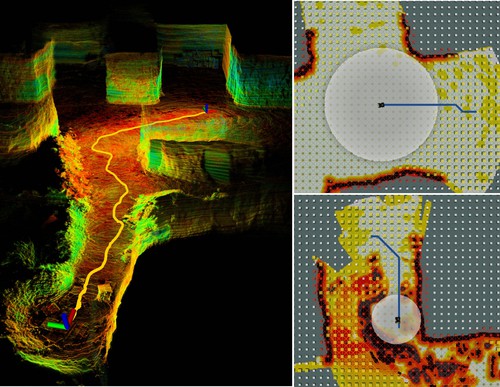}};
	    \begin{scope}[x={(image.south east)},y={(image.north west)}]

	    	\node [above left,align=right,white] at (0.18,0.09) {\textbf{A}};
	    	\node [above left,align=right,white] at (0.64,0.43) {\textbf{A}};
	    	\node [above left,align=right,white] at (0.465,0.715) {\textbf{B}};
	    	\node [above left,align=right,white] at (0.64,0.93) {\textbf{B}};

	    \end{scope}
	\end{tikzpicture}	
  \caption{Adaptive coverage range (translucent circle) and resulting exploratory path (blue) in a locally confined area (A) and a spacious area (B) during Husky's autonomous exploration of a limestone mine in Nicholasville, KY.} \label{fig:IRMs}
  \vspace{-5mm}
\end{figure}
The problem of finding the optimal sequence of sensing actions, or viewpoints, in order to maximize some task-specific information has been extensively studied, both in computer vision and robotics. 
In the robotics field, the problem of viewpoint selection is commonly motivated by tasks such as surveillance, object inspection, and exploration. While a variety of viewpoint selection algorithms have been proposed, we address those used to solve the exploration problem 
where policies are constructed in a receding horizon fashion as the robot gathers more sensory information about its environment. 

Viewpoint selection algorithms employ a sensor model to determine future sensing locations that maximize scene information.
In the context of exploration, these schemes often rely on identification of the boundary between unmapped and mapped space, regions termed \emph{frontiers}, and seek new robot poses that extend the boundary of mapped space \cite{yamauchi1997frontier}.
Traditional frontier-based approaches construct one-step lookahead policies that find the next most favorable sensing action, the quality of which is determined by the amount of unmapped area that can be visualized \cite{yamauchi1997frontier}, \cite{gonzalez2002navigation}.
Underpinning many approaches is the next-best-view planner (NBV) \cite{bircher2016receding}, where a rapidly exploring random tree is constructed. Each vertex represents a viewpoint, and the vertex that maximizes a utility function, weighing volumetric gain against path distance, is greedily selected as the next goal \cite{witting2018history}.
Dang et al. \cite{dang2020graph} extends this strategy by sampling a set of paths, and then selects the path which maximizes volumetric gain. 
While computationally efficient, NBV-based planners are greedy and therefore susceptible to local minima, leading to suboptimal decision making.
An accumulation of suboptimal local decisions can significantly reduce the amount of sensor information gathered over time.

In order to optimize viewpoint selection over a multi-step horizon, the exploration problem has been framed as a variant of the \emph{art gallery} problem \cite{ghosh2007visibility}. 
Here the objective is to find a minimal set of viewpoints that maximizes coverage of an area. 
A critical feature of this problem is the fact that the marginal benefit of selecting a new viewpoint decreases as the set of already selected viewpoints increases\,--\,a property known as \emph{submodularity}.
A greedy algorithm has been shown to provide a good approximation of the optimal solution to the submodular function maximization problem \cite{krause2014submodular}. 

Leveraging the effectiveness of greedy methods for submodular maximization, many have adopted a decoupled approach to the exploration problem \cite{cao2021tare}, \cite{heng2015efficient}, \cite{faigl2013determination}. 
First, sensing locations are selected using a greedy algorithm. Then a path through the locations is determined. 
For instance, in the work of Cao et al. \cite{cao2021tare}, a set of viewpoints is first sampled from a grid-based environment representation. Then viewpoints are selected in order of marginal coverage reward. To account for submodularity, the coverage rewards of the remaining viewpoints in the set are recomputed after each selection. 
The final ordering of viewpoints is determined by solving the standard \emph{traveling salesman problem}. While a decoupled approach provides a non-myopic solution in a computationally efficient manner, 
we contend that it can be sensitive to model uncertainty, which we discuss in Section \ref{sec:online}.

The main contribution of our work is a unified approach to the exploration problem that simultaneously considers environment coverage and robot traversability using a rollout-based search algorithm. 
The tractability of this approach relies on an approximation of the robot's coverage sensor model, which reduces planning time by adapting to the local environment.
We contend that our unified approach is more robust to real-world uncertainty than the widely-adopted decoupled method.

\section{Problem Definition}\label{sec:formulation}

\begin{figure}[t!]
  \centering
  \subfloat[Exact Coverage Range]{\label{fig:a}\includegraphics[width=.45\textwidth]{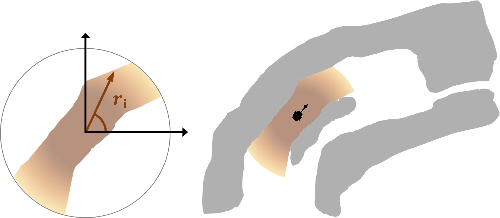}}
  \vspace{-4mm}
  \subfloat[Static Coverage Range]{\label{fig:b}\includegraphics[width=.45\textwidth]{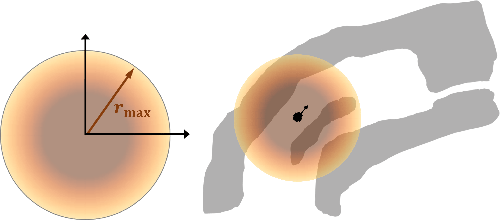}} 
  \vspace{-4mm}
  \subfloat[Adaptive Coverage Range (Proposed)]{\label{fig:a}\includegraphics[width=.45\textwidth]{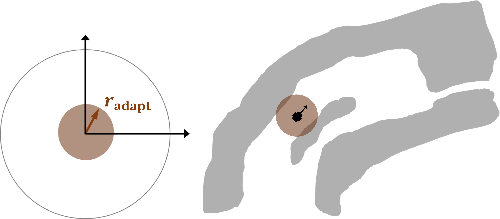}}

  \caption[The short caption]{Illustrative example of the effect of different coverage sensor models on exploration completeness: ``exact'' observation where the coverage range is based on ray-tracing (a), approximate observation where the coverage range is static (b), and our proposed approximate coverage sensor model where the range adapts to the local environment. While the exact model provides the best estimate of future coverage, it is computationally expensive and prevents proper investigation of the policy space during MCTS. Alternatively, while the static model is inexpensive, it overestimates the covered area. As a consequence, the passageway below the robot may not be explored since it provides erroneously low coverage reward.}
  \vspace{-4mm}
  \label{fig:overlap}
\end{figure}

Given a known environment represented by an abstract graph structure $G = (N, E)$, with free and occupied nodes $N_{free}\cup N_{occ} = N$, the coverage objective is to find a sequence of nodes $p = \{n_0, ... , n_{k-1}\} \subseteq N_{free}$ of arbitrary length $k$ such that the number of free nodes within an accumulated coverage sensor footprint $F$ is maximized, subject to a budget constraint:
\begin{equation}
\label{eq:modular}
\begin{aligned}
    &p^* = \argmax_{p} \; \sum_{n_i \in p} \, F(n_i), \\
    &\text{subject to} \quad a(p) \leq a_{\text{max}}, 
\end{aligned}
\end{equation}
\noindent where $a(p)$ is the path action cost, $a_{\text{max}}$ is a user-defined action cost budget, and the sensor footprint $F$ maps each node to a set of ``covered'' nodes: $F(n_i) = (n_{i_1}, n_{i_2},.., n_{i_j})$. 

Recall that the coverage problem exhibits submodularity; that is, the marginal benefit of appending the path with a node $n_2$ ``close'' to $n_1 \in p$ is less than that if $n_1 \not\in p$. 
To account for this \emph{diminishing returns} property, we define marginal coverage as the newly covered area, given all the previously visited nodes:
\begin{align}
    \label{eq:marg}
     \tilde{F}(n_i) = F(n_i \, | \, {n_0, .., n_{i-1}}).
\end{align}
\noindent Given this definition, we can recast Eq.~\ref{eq:modular} as a coverage problem with an additive reward structure: 
\begin{equation}
\label{eq:submodular}
\begin{aligned}
    &p^* = \argmax_{p} \; \sum_{n_i \in p} \, \tilde{F}(n_i), \\
    &\text{subject to} \quad a(p) \leq a_{\text{max}}.
\end{aligned}
\end{equation}

\noindent We refer to Eq.~\ref{eq:submodular} as our coverage problem for the remainder of the paper.

\section{Methodology}\label{sec:belief}

We model the coverage problem as a discrete-time sequential decision making process where the optimal policy is a sequence of 
actions chosen to maximize a cumulative coverage reward.
To find near-optimal policies in real-time, we employ a rollout-based search algorithm that estimates the value of an action sequence by simulating interactions between the robot and world. 
During a simulated episode, or rollout, the robot and world states evolve \emph{together} -- the robot executes an action and makes a coverage measurement of its environment, Eq.~(\ref{eq:marg}), which yields a subsequent robot-world state and reward. Thus, rollouts provide a method of solving the inherently submodular coverage problem in a unified manner, i.e. a policy is evaluated on both the accumulated marginal coverage reward and the path cost.  

We introduce our world representation (Section \ref{sec:world_rep}), and then model our coverage problem as a Markov decision process (Section \ref{sec:mdp}). To solve this problem in real-time on a computationally-constrained robot, we propose an effective approximation to the coverage sensor model, which significantly reduces rollout computation. As a result, we are able to construct high-quality coverage paths at a high planning rate (Section \ref{sec:online}).

\subsection{World Representation} \label{sec:world_rep}

\begin{figure}[t]
  \centering
  \subfloat{\label{fig:b}\includegraphics[width=0.95\columnwidth]{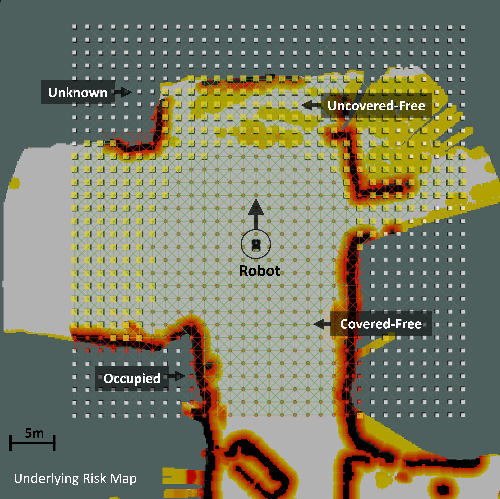}} 
  \caption[The short caption]{Information Roadmap (IRM) shown overlaid on the cost map. The IRM contains world coverage and traversability risk information. The goal of the coverage planner is to construct paths on the IRM that convert nodes from uncovered traversable (yellow) to covered traversable (brown). By constructing coverage paths in a receding-horizon fashion, the robot extends the boundaries of explored space.}
  \vspace{-3mm}
  \label{fig:local_irm_annotated}
\end{figure}

We represent the local environment around the robot by an information-rich graph structure called the Information Roadmap (IRM) \cite{sungpilgrim}, as shown in Fig. \ref{fig:local_irm_annotated}. The IRM is a fixed-size lattice graph $G = (N, E)$ with nodes $N$ and edges $E$. Nodes represent discrete areas in space, and edges represent actions.
We store two type of information in the IRM: \textit{(i)} the traversability risk of the world with respect to the robot's dynamic constraints, and \textit{(ii)} what parts of the environment have been observed, or \textit{covered}, by a task-specific coverage sensor.
The robot-centered, rolling window IRM is continuously updated with traversability and coverage information based on incoming sensor data.

To construct $G$, we uniformly sample nodes $n_i \in N$ in a neighborhood of the robot, and compute the traversability risk and coverage probability distribution over a discrete patch centered at each node, i.e., $p_r(n_{i})$ and $p_c(n_{i})$, which are stored as node properties. 
For scalability, we bin node traversability risk probabilities into three groups: \emph{occupied} $p_r(n_{i})=1$, \emph{unknown} $p_r(n_{i})=0.5$, and \emph{free} $p_r(n_{i})=0$.    
For an edge $e_{ij} \in E$, we compute and store the traversal distance $d_{ij}$ and traversal risk $\rho_{ij}$ between two connected nodes.

\subsection{Markov Decision Process} \label{sec:mdp}

A Markov decision process (MDP) is described as a tuple $\langle \mathbb{S}, \mathbb{A}, T, R \rangle$, where $\mathbb{S}$ is the set of joint robot-and-world states, and $\mathbb{A}$ is the set of robot actions. The motion model $T(s, a, s') = p(s'\,|\,s, a)$ defines the probability of being in state $s'$ after taking action $a$ in state $s$, and the reward function $R(s, a)$ returns the utility for executing action $a$ in state $s$. The objective is to find a mapping from states to actions, i.e. the policy $\pi$, that maximizes the expected sum of future reward. 

\ph{State}
The robot-world state is defined as $s = (q, W)$, where $q$ is the robot state and $W$ is the world state. We define $q$ and $W$ in terms of the IRM. The robot state $q=(n_q, \mu)$, where $n_q$ is the node closest to the robot's current location, and $\mu$ is the robot's heading direction, defined with respect to the lattice geometry. The world state is $W = G$, where $G$ is the IRM containing traversability risk and coverage world state estimates.

\ph{Action}
We define an action $a$ as the controlled robot traversal from node $n_i \in N$ to neighboring node $n_j \in N$, along an edge $e_{ij} \in E$. 
A node is directly connected to its eight neighbors, discretizing the valid action space for a single state into movement along the four cardinal/non-diagonal (N, E, S, W) and four intercardinal/diagonal (NE, SE, SW, NW) directions. We denote actions along the cardinal and intercardinal directions by
$a_{\sqrt{2}}$ and $a_{1}$, respectively.

\ph{Robot Dynamics}
We approximate the robot motion model $T(q, a, q')$ as deterministic. 
Given an action $a$ directing traversal of edge $e_{ij}$, the robot will reach node $n_j$ with probability 1. Actions that cause the robot to leave the bounds of $G$  or enter nodes that are unknown or occupied, $p_r(n_{i}) = 0.5$ or $1$, have no effect. 
Note that while don't explicitly model motion stochasticity, we account for it by planning at a high-rate in a receding-horizon fashion.

\ph{Probabilistic Coverage Sensor Model}
We model our coverage sensor as an omnidirectional range finder.
The robot covers nodes within its line-of-sight, computed using ray-tracing techniques on the traversal risk map $\{ p_r(n_{i}) \}$ in combination with sensor range constraints. 
To account for increasing ray sparsity in the radial direction, we compute the coverage probability for a node as a function of the robot-to-node distance. Given the robot node $n_q$, a node $n_i$ is covered with probability $P_{\text{cov}}(n_i | \; n_q)$. We heuristically model the coverage probability $P_{\text{cov}}$ as an S-shaped logistic function: 
\begin{align}
\label{eq:cov_prob}
    P_{\text{cov}}(n_i | \; n_q) = \frac{1}{1+e^{k \, (r_i-r_0)}},
\end{align}
\noindent where $r_{i}$ is the euclidean distance between the robot node $n_q$ and node $n_i$, and constants $r_0$ and $k$ are the sigmoid's midpoint and steepness, respectively. 
The coverage probability distribution over the radial distance from the center of the sensor in shown in Fig.~\ref{fig:overlap}.

\ph{World Transition Model}
We approximate the world transition function $T(W, a, W')$ as deterministic. 
Function $\text{\senseFunc}$ in Alg.~\ref{alg:sense} presents the process for updating the world coverage state based upon the the probabilistic coverage sensor model in Eq.~(\ref{eq:cov_prob}).
When integrating new sensor measurements, we assume independence and compute the maximum of the old and new coverage probability (Alg.~\ref{alg:sense}-line 5). 
This yields an optimistic estimate of coverage.

\begin{algorithm}[!t]
\small
\caption{World Coverage Update}
\label{alg:sense}
\hspace*{1em}\underbar{\textbf{Function \senseFunc}} \\
\hspace*{1em}\textbf{Input: }robot node $n_q$ \\
\hspace*{1em}\hspace*{2.8em} world state $G$ \\
\hspace*{1em}\hspace*{2.8em} maximum sensor range $r_{\text{max}}$ \\
\vspace*{-1em}
\begin{algorithmic}[1]
\FOR {all angles $\theta_k$ of range finder}
    \FOR {all nodes $n_i$ along ray from $n_q$ in direction $\theta_k$}
    \STATE Compute robot-to-node euclidean distance $r_{i}$
        \IF {$p_r(n_{i}) < \rho_{\text{max}}$ and $r_{i} < r_{\text{max}}$} 
        \STATE $p_c(n_{i})' \gets \text{max}\big[p_c(n_{i}), P_{\text{cov}}(n_i | \; n_q)\big]$ $\triangleright$ Eq.~(\ref{eq:cov_prob})
        \ELSE
        \STATE \textbf{break}
        \ENDIF
    \ENDFOR
\ENDFOR
\RETURN $\{ p_c(n_{i})' \}$ 
\end{algorithmic}
\end{algorithm}

\begin{figure}[!h]

\vspace{-4mm}

  \centering
  \subfloat[Coverage Probability (Continuous)]{\label{fig:a}\includegraphics[width=.22\textwidth]{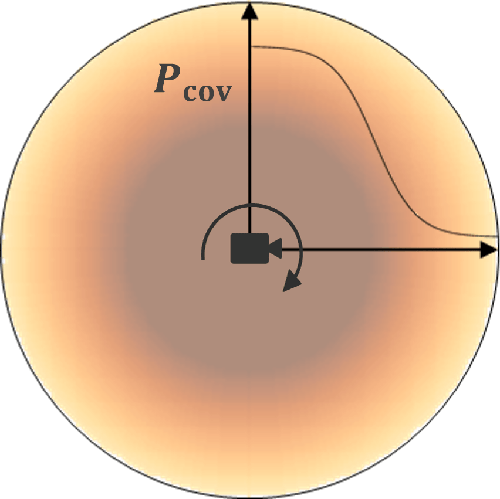}}
  \; \;
  \subfloat[Coverage Probability (Discretized)]{\label{fig:b}\includegraphics[width=.22\textwidth]{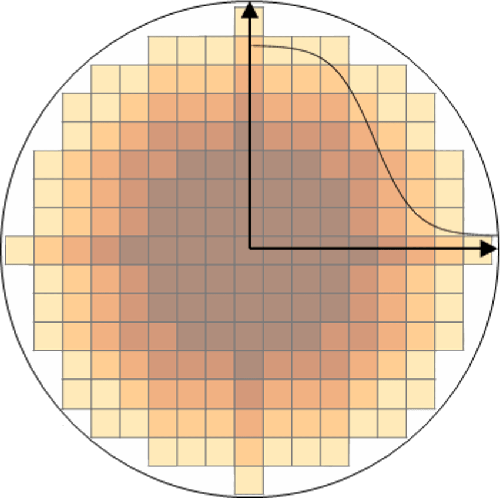}} 
  
  \subfloat[Non-Diagonal Action $a_{1}$]{\label{fig:a}\includegraphics[width=.235\textwidth]{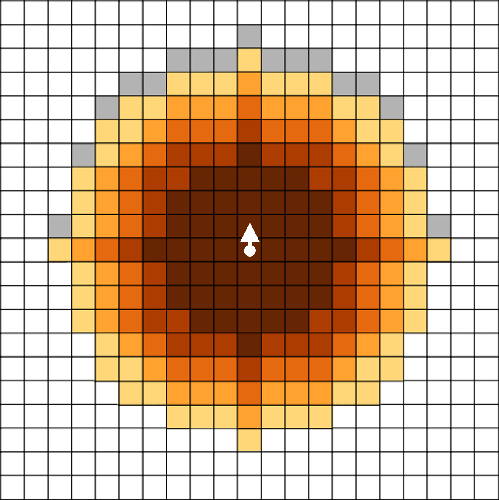}}
  \,
  \subfloat[Diagonal Action $a_{\sqrt{2}}$]{\label{fig:b}\includegraphics[width=.235\textwidth]{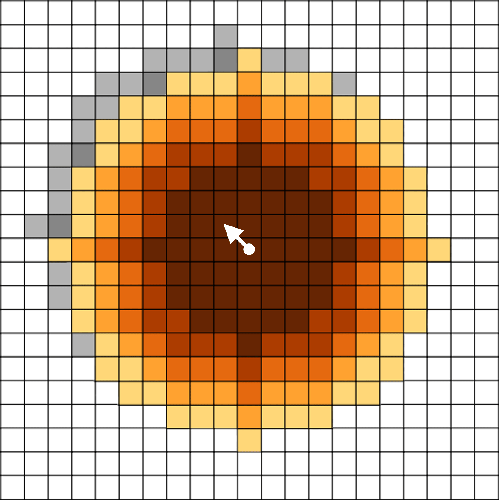}}     
  \caption[The short caption]{
  Our coverage sensor model, based on Eq.~(\ref{eq:cov_prob}), displayed over continuous space (a), and over the discretized lattice graph world representation (b). The diffused color map mimics the coverage probability curve-- darker shades indicate higher coverage probabilities. The marginal coverage after a non-diagonal action (c) and diagonal action (d) is represented by the shaded gray cells. Note that the ratio of marginal coverage to distance traveled over the lattice is not equivalent for non-diagonal and diagonal actions: $I(s^o, a_{\sqrt{2}})/d_{ij} \neq I(s^o, a_{1})/d_{ij}$, where $s^o$ indicates a risk-free world. We address this discrepancy with Eq.~(\ref{eq:distance_weight}).
  }
    \vspace{-4mm}
    
  \label{fig:adapt_cartoon}
\end{figure}

\ph{Reward Function}
We now redefine our marginal coverage from Eq.~(\ref{eq:marg}) to be the uncertainty reduction in the world coverage state induced by an action $a$:
\begin{align}
\label{eq:margCov}
    &I(s, a) = \sum_{n_i \in N} \beta \Big( p_c(n_{i} \, | \, a) - p_c(n_{i}) \Big),
\end{align}

\noindent where $\beta$ controls the reward received from covering a node based on its occupancy status.
Due to its sparsity, the IRM sometimes fails to identify nodes as occupied in high risk regions. For instance, in Fig. \ref{fig:local_irm_annotated}, the environment boundary is not fully represented by occupied nodes. 
To stay robust to this unreliable world model, we define the value of $\beta$ to be larger for nodes of known occupancy (occupied, uncovered-free, and covered-free), when compared to the value of $\beta$ for unknown nodes.
As a result, the constructed coverage paths are more likely to stay within the traversable space of the environment.

The reward function is defined as a weighted sum of marginal coverage and action penalties:
\begin{align}
\label{eq:reward}
    &R(s, a) = k_I \, I(s, a) - \big[ k_d \, d_{ij} + k_{\rho} \, \rho_{ij} + k_{\mu} \, \Delta_{\mu} \big], 
\end{align}
\noindent where $d_{ij}$ is the traversal distance, $\rho_{ij}$ traversal risk, and $\Delta_{\mu}$ is the cost of rotation due to the robot's non-holonomic constraints. Constants $k_I$, $k_d$, $k_\rho$, and $k_\mu$ weigh the importance of coverage, traversal distance, risk, and motion primitive history on the total reward.

Given a coverage sensor with a circular field-of-view, the uncovered area after a diagonal and non-diagonal action should scale equivalently with distance traveled. However, since Eq.~(\ref{eq:margCov}) is evaluated over a discretized space $G$, the ratio of marginal coverage to distance traveled is not equivalent for all actions on the lattice, as illustrated in Fig.~\ref{fig:adapt_cartoon}. Given this marginal coverage discrepancy between actions, we define $k_d$ as a function of coverage parameters in order to ensure non-diagonal ($a_1$) and diagonal actions ($a_{\sqrt{2}}$) are equally rewarding; that is, $R(s,a_1) = R(s,a_{\sqrt{2}})$ for the same $\rho_{ij}$ and $\Delta_{\mu}$. 
If $w$ is the width of a grid cell in $G$, then we define $k_d$ as: 
\begin{align}
\label{eq:distance_weight}
    k_d = \frac{k_I}{w} \cdot \frac{I(s^o, a_{\sqrt{2}}) - I(s^o, a_{1})}{(1-\sqrt{2})}
\end{align}
\noindent where state $s^o$ denotes a risk-free world where the only covered region is aligned with the robot's current sensor footprint.

\ph{Optimal Policy}
It is fundamentally infeasible to solve an unknown environment coverage problem over an infinite horizon since information about the world is incomplete, and often inaccurate, at runtime. Instead, in such domains, a Receding Horizon Planning (RHP) scheme has been widely adopted as the state-of-the-art \cite{bircher2016receding}. The optimal policy with RHP is:
\begin{align}
  \pi_{t:t+T}^*(s) &= \argmax_{\pi \in \Pi_{t:t+T}} \, \sum_{t'=t}^{t+T} \gamma^{t'-t} R(s_{t'}, \pi(s_{t'})),
  \label{eq:receding_objective_function}
\end{align}
where $T$ is a finite planning horizon for a planning episode at time $t$.
Given the policy from the last planning episode, only a part of the optimal policy, $\pi^*_{t:t+\Delta t}$ for $\Delta t \in (0, T]$, will be executed at runtime. A new planning episode will start at time $t+\Delta t$ with updated robot-world state.

\begin{algorithm}[t!]
\small
\caption{Coverage Planner}
\label{alg:planner}
\begin{algorithmic}
\STATE \underbar{\textbf{Function CoveragePlan}}
\REPEAT
    \item \textbf{Obtain: }state \state \\ %
    \hspace*{3.3em} pointcloud scan $\{ z_i \}$ \\
    \STATE \textbf{\#1 Generate Coverage Mask}
    \STATE Compute adaptive coverage range $r_{\text{adapt}}$ in Eq.~(\ref{eq:adapt})
    \STATE $\{m_{i} \} \gets \text{\senseFunc}(n_q, O,  r_{\text{adapt}})$ $\triangleright$ Alg.~\ref{alg:sense}
    \item \hspace*{2em} $\triangleright$ where $O \Rightarrow p_c(n_{i}) = p_r(n_{i}) = 0 \; \forall \; n_i \in N$
    \STATE \textbf{\#2 Find Planning Root}
    \STATE $n_{\tau}, \mu_{\tau} \gets \text{RootNode}(s, a^{-}_{1:N})$ $\triangleright$ see PLGRIM in \cite{sungpilgrim}
    \STATE $s \gets (n_{\tau}, \mu_{\tau}, G)$ $\triangleright$ update robot state to root parameters
    \STATE \textbf{\#3 Plan and Execute}
    \item $T_r \gets \textsc{MCTS}(s, \; \{m_{i} \})$
    \item Extract action sequence $a^*_{1:N}$ from $T_r$ %
    \STATE \textbf{\#4 Prep for Next Episode}
    \item $a^{-}_{1:N} \gets a_{1:N}$ 
\UNTIL timeout
\end{algorithmic}

\vspace*{0.2em}

\begin{algorithmic}
\STATE \underbar{\textbf{Function RootNode}}
\STATE \textbf{Input: }state \state \\
\hspace*{2.8em} previous action sequence $a^{-}_{1:N}$ \\
\vspace*{0.25em}
\STATE Extract path $a^{-}_{Q:N}$ $\triangleright$ $n_Q$ is path node closest to $n_q$
\STATE Initialize path risk $\rho_{\text{path}}$ and distance $d_{\text{path}}$ to 0
\FOR {action $e_{ij}$ in path $a^{-}_{Q:N}$}
\STATE $\rho_{\text{path}}$ += $\rho_{ij} / d_{ij}$; \; $d_{\text{path}}$ += $d_{ij}$
\IF{$\rho_{\text{path}} > \rho_{\text{max}}$ or $d_{\text{path}} > d_{\text{max}}$} 
\STATE Assign root node $n_{\tau} \gets n_i$
\STATE Find root orientation $\mu_{\tau}$ $\triangleright$ if $n_{\tau} = n_Q$, then $\mu_{\tau} \gets \mu$
\RETURN $n_{\tau}$, $\mu_{\tau}$
\ENDIF{}
\ENDFOR{}
\end{algorithmic}

\vspace*{0.2em}

\begin{algorithmic}
\STATE \underbar{\textbf{Function MCTS}}
\STATE \textbf{Input: }state \state \\
\hspace*{2.8em} coverage mask $\{ m_{i} \}$ \\
\vspace*{0.25em}
\STATE Initialize empty lookahead tree $T_r$
\REPEAT
    \item $T_r \gets \textsc{SIMULATE}\big(s; \; \mathcal{G}\big)$
    \item \hspace*{1em} $\triangleright$ estimated generative model $\mathcal{G}$ given by \item \hspace*{2em} $\text{Simulate}(s, \{ m_i \}; \, \pi_{rollout})$
\UNTIL{timeout}
\RETURN $T_r$
\end{algorithmic}

\vspace*{0.2em}

\hspace*{1em}\underbar{\textbf{Function Simulate}} \\
\hspace*{1em}\textbf{Input: }state \state \\
\hspace*{1em}\hspace*{3.1em}coverage mask $\{ m_{i} \}$ \\
\hspace*{1em}\hspace*{2.8em} policy $\pi$ 
\begin{algorithmic}
\STATE $n'_q, \, \mu' \gets \pi(n_q, \mu)$ %
\STATE $\{ p_c(n_{i})' \}  \gets \{ \text{max}\big[m_{i}, \; p_c(n_{i}) \big] \}$ $\triangleright$ fast coverage update
\STATE $r \gets R(s, a)$ $\triangleright$ Eq.~(\ref{eq:reward})
\RETURN $s', r$ %
\end{algorithmic}

\end{algorithm}

\subsection{Online Planning} \label{sec:online}

We now discuss our proposed online coverage planner algorithm, which runs in real-time on hardware. Alg.~\ref{alg:planner} presents the major components of the planner.

\ph{Search Algorithm}
In order to solve Eq.~(\ref{eq:receding_objective_function}), we use Monte Carlo tree search (MCTS) \cite{browne2012survey}.
Refer to Function MCTS in Alg.~\ref{alg:planner}.
During every planning episode, a lookahead tree, rooted in an initial robot-world state, is iteratively constructed by simulating action sequences using a random rollout policy $\pi_{rollout}$. 
During a single iteration, rollouts and tree expansion stop when a predefined depth, or our path budget, is reached.
Given a state $s$ and action $a$, a generative model $\mathcal{G}$ (i.e. the black box simulator of the MDP) provides a sample successor state $s'$ and reward $r$. 
Since we do not have access to the ground truth state of the environment, our generative model is an estimate based on the most recent robot sensor measurements used to construct the world representation $G$.
MCTS terminates after reaching a user-defined maximum number of simulations.

\ph{Action Sequence Extraction}
The action sequence with the highest estimated value is extracted from the lookahead tree (Alg.~\ref{alg:planner}\,--\,\#3). Then the first $N$ actions from that sequence, $a^*_{1:N}$, is sent to the robot for execution. 
The number of actions $N$ is defined such that $R(s_i,a_i)>\gamma \; \forall \; i \in \{ 1:N \}$, where $\gamma$ is an empirically selected one-step reward lower bound. This cropping of the action sequence is critical to global exploration performance; it ensures the local coverage path uncovers ``enough'' area to justify the path travel cost. If $a^*_{1:N}$ is empty, then a global planner takes control and guides the robot to areas with high expected information gain.

\begin{figure*}[!t]
  \centering
  \subfloat[t = 00:00]{\label{fig:a}\includegraphics[width=.24\textwidth]{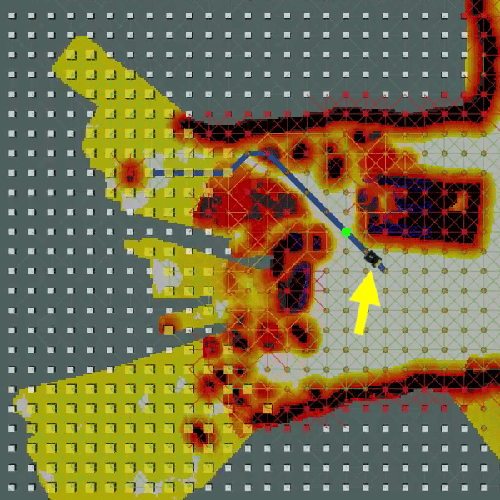}}
  \hspace{0.00005em}
  \subfloat[t = 00:30]{\label{fig:b}\includegraphics[width=.24\textwidth]{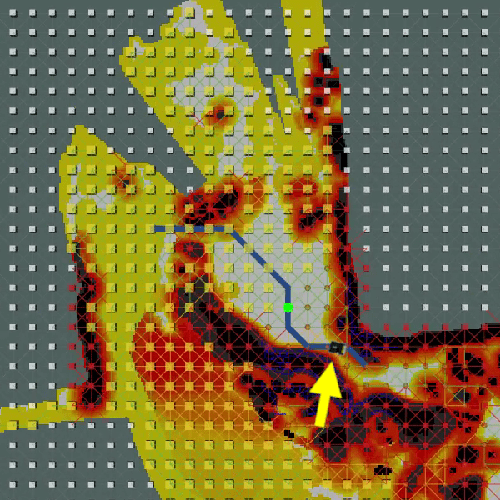}}
  \hspace{0.00005em}
  \subfloat[t = 00:31]{\label{fig:a}\includegraphics[width=.24\textwidth]{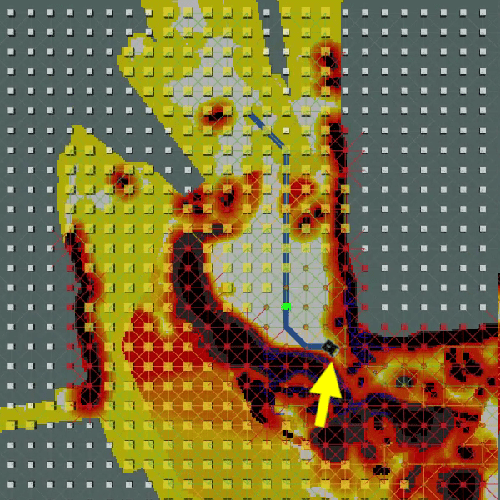}}
  \hspace{0.00005em}
  \subfloat[t = 01:46]{\label{fig:b}\includegraphics[width=.24\textwidth]{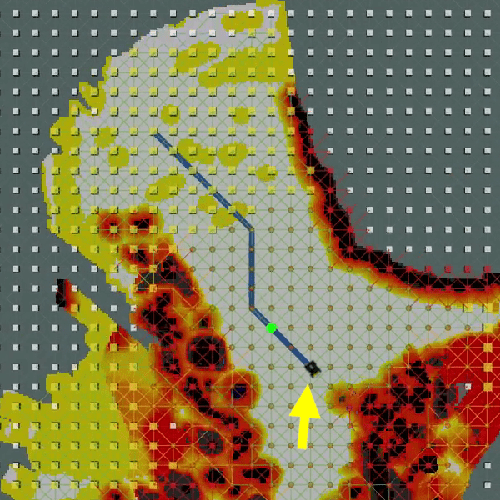}}   
  \caption[The short caption]{Snapshots of robot's navigation through rocks and debris during its exploration of a limestone mine. The coverage path (blue) and the planning root node (green circle) are shown. Note that (b) and (c) are from consecutive planning episodes as the robot turns a corner, receives new sensor information, and updates the world risk state. 
  The policy constructed in (c) is evaluated in (b)'s updated world estimate. The root node location is based on this evaluation (See Function RootNode in Alg. \ref{alg:planner}).}
  \label{fig:rock_pile}
  
  \vspace{-4mm}
\end{figure*}

\ph{Planning Root Update}
At the end of every planning episode, $a^*_{1:N}$ is stored  and then used to update the root of the lookahead tree during the subsequent episode (Alg.~\ref{alg:planner}\,--\,\#2). Our root update approach is based on a receding-horizon policy reconciliation method proposed by \cite{sungpilgrim}.
Fig. \ref{fig:rock_pile} demonstrates the effectiveness of this root update method.

\ph{Adaptive Coverage Range} 
While MCTS is an anytime algorithm, meaning construction of the tree can terminate at any point and a solution will be recovered, it only converges to the optimal solution with a sufficient number of simulations.
Although it may be infeasible to reach the optimal solution given time constraints, estimates of the action values become increasingly more reliable with more simulations, leading to a higher quality coverage path. 
In order to find quality solutions at high planning rates, a real-time system 
must find a good balance between the fidelity of a simulation (e.g. how accurately we model the coverage observation) and the number of simulations. 

To maximize the number of simulations within a suitable planning time, we propose an approximation of the coverage model that reduces the time complexity of the generative model $\mathcal{G}$. Our approximate world coverage update obviates the need for expensive ray-tracing operations in Alg.~\ref{alg:sense}. 
First, we estimate the spaciousness $r_{\text{spac}}$ of the local environment \cite{chen2022direct}. Then we adapt the distance at which a range-finder coverage measurement is performed based on $r_{\text{spac}}$. 
We denote this adaptive coverage distance by $r_{\text{adapt}}$.
See Fig. \ref{fig:overlap} as an example of our adaptive coverage range approach.

Given a range-finder 3D pointcloud scan $\{ z_i \}$ where $z_i$ is the point at which a ray intersects an obstacle, we compute spaciousness as:
\begin{align}
\label{eq:spaciousness}
    r_{\text{spac}}= f\big(\text{median} \{ d(z_i) \} \big),
\end{align}
\noindent where $d(z_i)$ is the euclidean distance between the range-finder origin and a ray intersection-point $z_i$, 
and $f$ is a low-pass filter: $f(x_t)=\alpha_1 \, f(x_{t-1}) + \alpha_2 \, x_{t}$ with constants $\alpha_1=0.95$ and $\alpha_2=0.05$.
The median is robust to outliers in a potentially noisy pointcloud, and gives a notion of the current scale of the local environment around the robot.
Then, given $r_{\text{spac}}$, we compute $r_{\text{adapt}}$ as
\begin{align}
    \label{eq:adapt}
    r_{\text{adapt}} = 
    \begin{cases}
    \alpha \cdot r_{\text{spac}}, \; \; \; \text{if} \; \; r_{\text{spac}} \leq \frac{r_{\text{max}}}{\alpha} 
    \\[2pt]
    r_{\text{max}}, \; \; \; \; \; \; \; \; \text{otherwise}, 
    \end{cases}
\end{align}
\noindent where $\alpha$ is an empirically tuned scaling constant, and $r_{\text{max}}$ is our model-defined maximum sensor range. Equipped with $r_{\text{adapt}}$, we generate a probabilistic coverage mask $\{m_{i} \}$, detailed in Alg.~\ref{alg:planner}\,--\,\#1. The mask serves as an input to the generative model Function Simulate in Alg.~\ref{alg:planner}, which updates the world coverage state using inexpensive matrix operations.

\begin{figure}[!t]
  \centering
  \subfloat[Unified Coverage Planner (Proposed)]{\label{fig:b}\includegraphics[width=1.0\columnwidth]{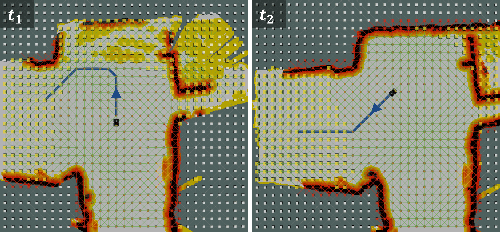}} 
  \vspace{-2mm}
  \subfloat[Decoupled Coverage Planner]{\label{fig:b}\includegraphics[width=1.0\columnwidth]{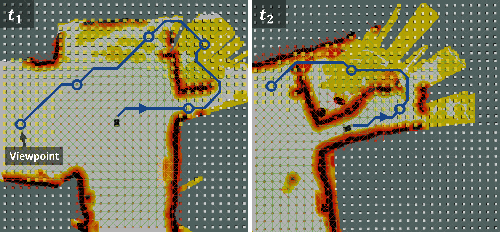}} 
  \caption[The short caption]{
    For two planning episodes at $t_1$ and $t_2$ during Husky's autonomous exploration of a real-world mine, we show the coverage path constructed using our proposed unified approach (a) and the commonly-adopted decoupled approach (b) for solving the submodular coverage problem.
    In (a), the robot collects the remaining coverage reward at the end of the passage, before continuing to the large, unexplored passage to the left. 
    In (b) at snapshot $t_1$, the robot incorrectly detects openings at the end of the passage due to bad sensor measurements and selects a set of viewpoints accordingly. The shortest path through the poorly-selected viewpoints guides the robot through a narrow passage to the right, which is both riskier and less rewarding than the passage to the left. 
    }
  \label{fig:local_irm}
 \vspace{-4mm}
\end{figure}

\ph{Discussion}
A decoupled approach to the coverage planning problem leverages a greedy algorithm for non-myopic viewpoint selection, as detailed in Section \ref{sec:relatedwork}. This approximation relies on the fact that selecting more viewpoints never reduces the total coverage reward, since Eq.~\ref{eq:submodular} exhibits monotonicity \cite{krause2014submodular, roberts2017submodular}. 
While true in theory, this conjecture falters in a real-world exploration domain where the robot only has partial information about the world.
In this setting, the inclusion of risky or low quality viewpoints, i.e., those evaluated using unreliable world estimates, can have adverse effects on the final policy and the robot's ability to collect coverage reward over an exploration mission. 
More concretely, the policy constructed by a decoupled approach does not consider that: (\emph{i}) the robot may fail during execution of the path, (\emph{ii}) world coverage and traversability estimates become increasingly unreliable with increasing distance from the robot, and (\emph{iii}) the world model (i.e. Local IRM) changes dynamically as the robot uncovers and maps new regions. 

In order to address the aforementioned issues, 
the proposed approach to the coverage planning problem exhibits the following properties that make it suitable for a real-world exploration domain.
\begin{enumerate}[label={\arabic*)}]

  \item \textit{Viewpoint Selectiveness}:
    A policy is evaluated by computing the marginal coverage reward and path cost for each successive action, or viewpoint, in the policy (Eq.~\ref{eq:reward}).
    Understanding coverage interdependency between successive viewpoints lifts the burden of 
    needing to fully cover the current graph with a single policy\,--\,an unproductive and potentially harmful ambition in the presence of uncertainty. 
    As a result, viewpoints that do not provide sufficient coverage utility within a time-budget, or jeopardize the robot's safety,
    can be discounted from the final policy, while still preserving MCTS near-optimality. 

  \item \textit{Robustness to Uncertainty}:
    The lookahead tree is rooted at (or very near to) the robot's current location.
    Hence, MCTS visits nodes close to the robot more frequently, effectively focusing its search time in areas of the environment where world coverage and traversability risk estimates are more reliable. 
    Moreover, due to a discount factor in the problem objective Eq.~(\ref{eq:receding_objective_function}), policies that shift coverage reward earlier in time are more rewarding.
    By incorporating this near-sighted incentive, the robot accounts for stochasticity in sensing and motion control, 
    as well as the fact that the world model will evolve as undetected areas are exposed. 
  
\end{enumerate}

Fig.~\ref{fig:local_irm} compares paths constructed by our approach and a decoupled approach during a real-world exploration mission. 
Recall that the decoupled approach greedily selects viewpoints in order of highest marginal coverage reward. Therefore, rather than discounting viewpoints far from the robot where world estimates are poor, the decoupled approach actually prioritizes distant points since there is less sensor overlap at these locations with the robot's current field-of-view. The path planner is then ``locked into'' these viewpoints, and optimistically reasons over this potentially unreliable search space.

\section{Experimental Results}\label{sec:experiments}

In order to evaluate our proposed approach, we performed simulation studies and real-world experiments with a four-wheeled vehicle (Clearpath Robotics Husky robot) and quadruped (Boston Dynamics Spot robot). Both robots are equipped with custom sensing and computing systems~\cite{Otsu2020,AliNeBula21,AutoSpot}. The entire autonomy stack runs in real-time on an Intel Core i7 processor with 32 GB of RAM. The stack relies on a multi-sensor fusion framework, the core of which is 3D point cloud data provided by LiDAR range sensors~\cite{Ebadi2020}. 
During testing, the proposed (or comparative baseline) approach was integrated as the local planner within the hierarchical planning framework PLGRIM \cite{sungpilgrim}. 

\subsection{Simulation Evaluation}

\begin{figure}[!b]
  \centering
  \subfloat{\label{fig:b}\includegraphics[width=1.0\columnwidth]{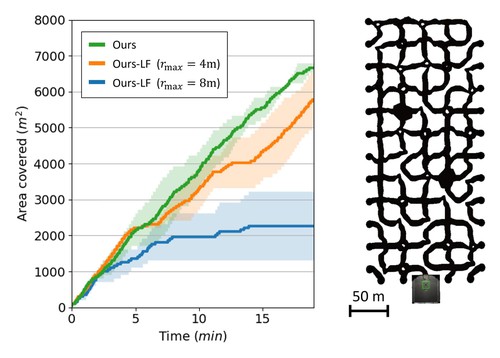}}
  \caption[The short caption]{Results from simulated exploration runs in the simulated maze (shown at top right). We define our coverage metric to be the accumulated area within an 8m radius of the robot. Each curve is the average of 2 runs. }
  \label{fig:local_maze_sim}
\end{figure}

We evaluated the proposed planner against the baseline planner \emph{Ours-LF}: the proposed rollout-based method with a low-fidelity coverage sensor model, i.e. non-probabilistic and static coverage range $r_{\text{max}}$.
All tests were performed in a simulated maze environment, as shown in Fig. \ref{fig:local_maze_sim}.
The maze consists of a large irregular network of large spaces and narrow passages, many of which are connected by sharp bends. 
This geometry exposes the weaknesses of a rollout-based planner where the coverage sensor model does not effectively approximate the actual range finder sensor.
The long-range Ours-LF planner ($r_{max}$ = 8m) overestimates the coverage sensor range and, therefore, fails to detect openings at the sharp bends. As a result, large swaths of the environment are not exposed, and the robot terminates exploration early. 
Alternatively, the short-range Ours-LF planner ($r_{max}$ = 4m) performs significantly better since it can expose and explore all narrow passages. 
However, since it underestimates the coverage sensor range, it finds redundant trajectories in the large spaces, which contributes to a slight degradation in performance.

Our proposed solution can handle all settings, since it neither over- or under-estimates the true coverage range in exchange for reducing computation. Moreover, since the model is probabilistic, it inherently adjusts its coverage density to the local environment. As a consequence, when the robot approaches a sharp bend, it travels deep enough to ``see'' uncovered space around the corner, which is critical to exposing the entire environment.

\subsection{Real-World Evaluation}
\begin{figure}[h!]
  \centering
  \subfloat{\label{fig:b}\includegraphics[width=1.0\columnwidth]{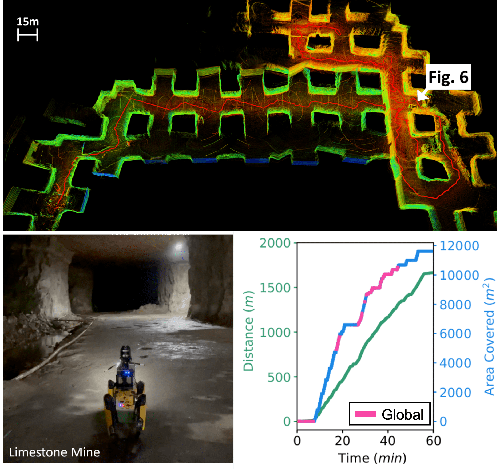}}
  \caption[The short caption]{Proposed coverage planner's navigation of a limestone mine during a 60 min. exploration mission. The coverage paths down the main corridor exhibit a wave-like shape. When the robot encounters a junction, it moves toward the corner in order to maximize coverage of both branches, and then re-aligns with the centerline of the main corridor. In the exploration metric (bottom right), pink denotes the time intervals where the the PLGRIM global planner is directly guiding the robot. The starting location of snapshot (a) in Fig. \ref{fig:rock_pile} is indicated. }
  \label{fig:limestone_mine}
\end{figure}

Our solution was extensively tested on physical robots in real-world environments. In particular, we present results from the exploration of a limestone mine (Figs. \ref{fig:limestone_mine} and \ref{fig:rock_pile}) in the Kentucky Underground, Nicholasville, KY.

\section{Conclusion}\label{sec:conclusion}
We present an approach for solving the coverage problem for time-constrained autonomous exploration of unknown environments. We solve this problem, which is submodular in nature, using a unified rollout-based search algorithm. 
This formulation allows us to evaluate the effects of the robot's actions on future world coverage states, while simultaneously accounting for traversability risk and the dynamic constraints of the robot.
In order to adequately investigate the search space, we reduce rollout computation using an effective approximation to the coverage sensor model which adapts the coverage range to the local environment. 
As a result, we can solve the submodular coverage problem in a unified manner, which we contend is more robust to real-world uncertainty than decoupled approaches.

\section*{Acknowledgments}
We acknowledge Mamoru Sobue, Harrison Delecki, Oriana Peltzer and all team members of Team CoSTAR for the DARPA Subterranean Challenge, and the resource staff at the Kentucky Underground facility, Nicholasville, KY.

\bibliographystyle{IEEEtran}
\bibliography{references}

\begin{thebibliography}{10}
\providecommand{\url}[1]{#1}
\csname url@samestyle\endcsname
\providecommand{\newblock}{\relax}
\providecommand{\bibinfo}[2]{#2}
\providecommand{\BIBentrySTDinterwordspacing}{\spaceskip=0pt\relax}
\providecommand{\BIBentryALTinterwordstretchfactor}{4}
\providecommand{\BIBentryALTinterwordspacing}{\spaceskip=\fontdimen2\font plus
\BIBentryALTinterwordstretchfactor\fontdimen3\font minus
  \fontdimen4\font\relax}
\providecommand{\BIBforeignlanguage}[2]{{%
\expandafter\ifx\csname l@#1\endcsname\relax
\typeout{** WARNING: IEEEtran.bst: No hyphenation pattern has been}%
\typeout{** loaded for the language `#1'. Using the pattern for}%
\typeout{** the default language instead.}%
\else
\language=\csname l@#1\endcsname
\fi
#2}}
\providecommand{\BIBdecl}{\relax}
\BIBdecl

\bibitem{sungpilgrim}
S.-K. {Kim$*$}, A.~{Bouman$*$}, G.~Salhotra \emph{et~al.}, ``{PLGRIM}:
  Hierarchical value learning for large-scale exploration in unknown
  environments,'' in \emph{International Conference on Automated Planning and
  Scheduling (ICAPS)}, vol.~31, 2021, pp. 652--662.

\bibitem{heng2015efficient}
L.~Heng, A.~Gotovos, A.~Krause, and M.~Pollefeys, ``Efficient visual
  exploration and coverage with a micro aerial vehicle in unknown
  environments,'' in \emph{IEEE International Conference on Robotics and
  Automation (ICRA)}.\hskip 1em plus 0.5em minus 0.4em\relax IEEE, 2015, pp.
  1071--1078.

\bibitem{AliNeBula21}
A.~Agha-mohammadi and {et al.}, ``{NeBula}: Quest for robotic autonomy in
  challenging environments; {TEAM CoSTAR} at the {DARPA} subterranean
  challenge,'' \emph{Journal of Field Robotics}, 2021.

\bibitem{yamauchi1997frontier}
B.~Yamauchi, ``A frontier-based approach for autonomous exploration,'' in
  \emph{IEEE International Symposium on Computational Intelligence in Robotics
  and Automation}, 1997, pp. 146--151.

\bibitem{gonzalez2002navigation}
H.~H. Gonz{\'a}lez-Banos and J.-C. Latombe, ``Navigation strategies for
  exploring indoor environments,'' \emph{International Journal of Robotics
  Research}, vol.~21, no. 10-11, pp. 829--848, 2002.

\bibitem{bircher2016receding}
A.~Bircher, M.~Kamel, K.~Alexis \emph{et~al.}, ``Receding horizon
  ``next-best-view'' planner for {3D} exploration,'' in \emph{icra}, 2016, pp.
  1462--1468.

\bibitem{witting2018history}
C.~Witting, M.~Fehr, R.~B{\"a}hnemann \emph{et~al.}, ``History-aware autonomous
  exploration in confined environments using mavs,'' in \emph{2018 IEEE/RSJ
  International Conference on Intelligent Robots and Systems (IROS)}.\hskip 1em
  plus 0.5em minus 0.4em\relax IEEE, 2018, pp. 1--9.

\bibitem{dang2020graph}
T.~Dang, M.~Tranzatto, S.~Khattak \emph{et~al.}, ``Graph-based subterranean
  exploration path planning using aerial and legged robots,'' \emph{Journal of
  Field Robotics}, vol.~37, no.~8, pp. 1363--1388, 2020.

\bibitem{ghosh2007visibility}
S.~K. Ghosh, \emph{Visibility algorithms in the plane}.\hskip 1em plus 0.5em
  minus 0.4em\relax Cambridge University Press, 2007.

\bibitem{krause2014submodular}
A.~Krause and D.~Golovin, ``Submodular function maximization.''
  \emph{Tractability}, vol.~3, pp. 71--104, 2014.

\bibitem{cao2021tare}
C.~Cao, H.~Zhu, H.~Choset, and J.~Zhang, ``Tare: A hierarchical framework for
  efficiently exploring complex 3d environments,'' in \emph{Robotics: Science
  and Systems Conference (RSS), Virtual}, 2021.

\bibitem{faigl2013determination}
J.~Faigl and M.~Kulich, ``On determination of goal candidates in frontier-based
  multi-robot exploration,'' in \emph{2013 European Conference on Mobile
  Robots}.\hskip 1em plus 0.5em minus 0.4em\relax IEEE, 2013, pp. 210--215.

\bibitem{browne2012survey}
C.~B. Browne, E.~Powley, D.~Whitehouse \emph{et~al.}, ``A survey of \text{Monte
  Carlo Tree Search} methods,'' \emph{IEEE Transactions on Computational
  Intelligence and AI in games}, vol.~4, no.~1, pp. 1--43, 2012.

\bibitem{chen2022direct}
K.~Chen, B.~T. Lopez, A.-a. Agha-mohammadi, and A.~Mehta, ``Direct lidar
  odometry: Fast localization with dense point clouds,'' \emph{IEEE Robotics
  and Automation Letters}, vol.~7, no.~2, pp. 2000--2007, 2022.

\bibitem{roberts2017submodular}
M.~Roberts, D.~Dey, A.~Truong \emph{et~al.}, ``Submodular trajectory
  optimization for aerial 3d scanning,'' in \emph{Proceedings of the IEEE
  International Conference on Computer Vision}, 2017, pp. 5324--5333.

\bibitem{Otsu2020}
K.~Otsu, S.~Tepsuporn, R.~Thakker \emph{et~al.}, ``{Supervised autonomy for
  communication-degraded subterranean exploration by a robot team},'' in
  \emph{IEEE Aerospace Conference}, 2020.

\bibitem{AutoSpot}
A.~{Bouman$*$}, M.~{Ginting$*$}, N.~{Alatur$*$} \emph{et~al.}, ``{Autonomous
  Spot: Long-Range Autonomous Exploration of Extreme Environments with Legged
  Locomotion},'' in \emph{iros}, 2020.

\bibitem{Ebadi2020}
K.~Ebadi, Y.~Chang, M.~Palieri \emph{et~al.}, ``{LAMP}: Large-scale autonomous
  mapping and positioning for exploration of perceptually-degraded subterranean
  environments,'' in \emph{icra}, 2020.

\end{thebibliography}

\end{document}